\documentclass{article}

\usepackage[english]{babel}

\usepackage[letterpaper,top=2cm,bottom=2cm,left=3cm,right=3cm,marginparwidth=1.75cm]{geometry}

\usepackage{amsmath}
\usepackage{graphicx}
\usepackage[colorlinks=true, allcolors=blue]{hyperref}
\usepackage{hyperref}
\usepackage{url}
\usepackage{array}
\usepackage{algorithm}
\usepackage{algpseudocode}
\graphicspath{ {./Images/} }
\usepackage[table]{xcolor}
\usepackage[aboveskip=1pt,labelfont=bf,labelsep=period,justification=raggedright,singlelinecheck=off]{caption}
\usepackage{lastpage,fancyhdr,graphicx}
\usepackage{epstopdf}
\usepackage{amsmath,amssymb}
\usepackage{textcomp,marvosym}
\usepackage{changepage}
\title{CleftGAN: Adapting A Style-Based Generative Adversarial Network To Create Images Depicting Cleft Lip Deformity}
\date{\vspace{-5ex}}

\begin{document}
\maketitle

\begin{flushleft}
\author{Abdullah Hayajneh\textsuperscript{1},
Erchin Serpedin\textsuperscript{1},
Mohammad Shaqfeh\textsuperscript{2},
Graeme Glass\textsuperscript{3}
Mitchell A. Stotland\textsuperscript{4}
\\
\textbf{1}  Electrical and Computer Engineering Department, Texas A\&M University, College Station, TX, USA
\\
\textbf{2} Electrical and Computer Engineering Program, Texas A\&M University, Doha, Qatar
\\
\textbf{3} Division of Plastic, Craniofacial and Hand Surgery, Sidra Medicine.
\\
\textbf{4} Division of Plastic, Craniofacial and Hand Surgery, Sidra Medicine, and Weill Cornell Medical College, Doha, Qatar
\\

}
\end{flushleft}
\begin{abstract}
A major obstacle when attempting to train a machine learning system to evaluate facial clefts is the scarcity of large datasets of high-quality, ethics board-approved patient images. In response, we have built a deep learning-based cleft lip generator designed to produce an almost unlimited number of artificial images exhibiting high-fidelity facsimiles of cleft lip with wide variation. 
We undertook a transfer learning protocol testing different versions of StyleGAN-ADA (a generative adversarial network image generator incorporating adaptive data augmentation (ADA)) as the base model. Training images depicting a variety of cleft deformities were pre-processed to adjust for rotation, scaling, color adjustment and background blurring. The ADA modification of the primary algorithm permitted construction of our new generative model while requiring input of a relatively small number of training images. Adversarial training was carried out using 514 unique frontal photographs of cleft-affected faces to adapt a pre-trained model based on 70,000 normal faces. The Frechet Inception Distance (FID) was used to measure the similarity of the newly generated facial images to the cleft training dataset, while
Perceptual Path Length (PPL) and the novel Divergence Index of Severity Histograms (DISH) measures were also used to assess the performance of the image generator that we dub CleftGAN. 
CleftGAN demonstrates an ability to automatically generate vast numbers of unique faces depicting a wide range of cleft lip deformity including a variation of ethnic/racial background. We found that StyleGAN3 with translation invariance (StyleGAN3-t) performed optimally as a base model. Generated images achieved a low FID reflecting a close similarity to our training input dataset of genuine cleft images. Low PPL and DISH measures reflected a smooth and semantically valid interpolation of images through the transfer learning process and a similar distribution of severity in the training and generated images, respectively. 
We introduce CleftGAN, a novel instrument capable of generating an almost boundless number of realistic facial images depicting cleft lip. This tool promises to become a valuable resource for the development of machine learning models to objectively evaluate facial form and the outcomes of surgical reconstruction. CleftGAN has been made available to public at: \url{https://github.com/abdullah-tamu/CleftGAN}

\end{abstract}

\begin{figure}[h]
\begin{center}
\framebox{
\includegraphics[width=12cm]{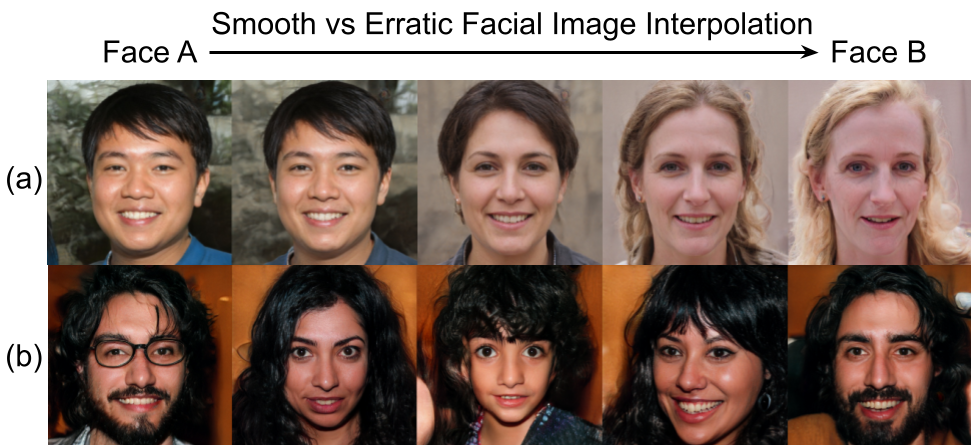}}
\end{center}
\caption{\label{img:StyleGAN_smoothness}Row (a): Example of smooth, step-wise, semantic interpolation between Face A and Face B. Row (b): Example of erratic, discontinuous, intermediate transitions between Face A and Face B. All images generated from the StyleGAN2 facial generator \cite{karras2019style}.}
\end{figure}


\section*{Introduction}
The coupling of image processing with artificial intelligence methods may offer an ideal pathway towards objective, clinical evaluation of the human face. Management of congenital and acquired forms of facial difference would benefit considerably from a universally accessible and standardized form of measurement.  Cleft lip deformity (with or without a cleft palate) occurs in approximately 1/1000 live births in the United States \cite{mai2019national} and is the focus of this current research protocol. Patients affected by cleft lip commonly undergo multiple surgical interventions over the course of their childhood in an effort to diminish evidence of abnormality. Decision-making regarding timing, extent, and outcome of surgical reconstruction is primarily based on subjective assessment on the part of the patient, parents, and provider. In order to train objective machine learning-based assessment models typically requires large training datasets on the order of tens of thousands of images. Obtaining such vast collections of high-quality, ethics board-approved and consented images of children with cleft deformity is a formidable challenge for any institution or research consortium due to many parents’ reluctance to have their children’s faces studied, shared, and published.
Existing facial datasets used in the domain of machine learning research are numerous and include Celeb-A \cite{liu2018large}, Tufts Face Dataset \cite{panetta2018comprehensive}, VGG-FACE2 \cite{cao2018vggface2}, Google Facial Expression Comparison Dataset \cite{vemulapalli2019compact}, and many others \cite{BestFacialDatasets2022,Top10FaceDatasets}. To our knowledge, however, there are no facial datasets available that include images depicting cleft lip deformity. The preponderance of facial image processing research uses collections of normal faces in order to train facial recognition systems for the purposes of identification verification, security, or morphologic diagnosis \cite{raji2021face,li2020review,hallgrimsson2020automated}. One critical goal of our ongoing research, however, is to develop a computer model that will objectively appraise facial cleft morphology in a manner similar to human judgment. Creating a reliable and practical scale with which to gauge the human face would facilitate (i) impartial self-assessment of clinical outcome on the part of the surgeon, (ii) meaningful discussion with patient or parent, (iii) outcome comparison between different surgical techniques and surgeons, (iv) enhanced surgical planning and education, (v) research on factors associated with cleft severity, and (vi) explicit characterisation of the clinical needs and benefits of surgery for third-party payers. Construction of such a model, however, requires large numbers of images of affected faces.

A generative adversarial network (GAN) \cite{goodfellow2020generative} capable of producing cleft images would therefore support such work in an important manner. A GAN consists of two networks: a generator G and a discriminator D. The task of the generator is to produce realistic facsimile images. The discriminator - trained on a dataset of real images - determines the likelihood that the generator’s images are real or not. In an iterative process for each image, the discriminator feeds back an error signal, directing the generator to gradually improve the authenticity of its fabricated images. One example of such a system is the StyleGAN \cite{karras2019style}, a computer model introduced in December 2018 that has received a significant amount of popular attention (a Google search on June 1, 2023 yielded more than 430,000 Google results). An updated iteration, StyleGAN2 \cite{karras2020analyzing}, powers the open-source website “thispersondoesnotexist.com” which generates an enormous array of highly realistic human faces across a broad spectrum of age, gender, and race. In the current research protocol we carried out a transfer learning protocol testing three updated versions of the system: StyleGAN2-ADA \cite{karras2020training}) which is able to adaptively incorporate different types of data augmentations into its training flow, as well as StyleGAN3-t (better translation equivariance) and StyleGAN3-r (better rotation equivariance) \cite{karras2021alias}. 

An ideal cleft generator should have the following characteristics: (i) consistently provides a diverse set of high quality and genuine looking faces with different types of cleft anomalies, (ii) generates an almost limitless number of unique faces. If based on a large enough dataset, the permutations and combinations of image elements that such a GAN is able to recombine into novel arrangements will be immense, yielding a variation in ethnicity, facial expression, lighting, pose, orientation, size, etc., and (iii) allows for a smooth and semantically valid interpolation between the generated faces as demonstrated in figure \ref{img:StyleGAN_smoothness},  where a smooth path from image A to B is shown, whereas an inefficient and indirect path between A to B is generated in row b.

\section{Materials and Methods}

\subsection{Dataset Collection}
A total of 514 facial images were used to train our machine learning model. These images were obtained from two sources:
\begin{itemize}
    \item Patient images were collected with signed informed consent from the Division of Plastic, Craniofacial and Hand Surgery at Sidra Medicine, Doha, Qatar within an IRB-approved protocol. These images were all neutral frontal poses obtained using a Nikon D610 camera (Nikon, Inc., Tokyo, Japan) at a resolution of 300 dpi and with dimensions of 6016 $\times$ 4016 pixels.
    \item Open source images depicting various types of cleft lip deformity were crawled from the internet from an array of different sites. These images were variable in terms of pose, orientation, lighting, and resolution, and ranged from  $372 \times 348$  to  $5726 \times 4295$  pixels in size. In general, the  open source images were of poorer quality and contained more noise than the clinical photos. 
\end{itemize}

Our transfer learning protocols were executed in stages with progressively larger image datasets: (1) a sample set of 250 open source images; (2) a combination of 391 open source and 59 clinical images for a total of 450; and (3) a final testing phase with a total of 391 open source and 123 clinical photos = 514 facial images.

\subsection{Dataset Preprocessing}
Preprocessing steps were necessary in order to sync our image set with the StyleGAN pretrained networks which model faces with specific pose, orientation, and face-to-background ratio. All images were thus uniformly scaled to $1024 \times 1024$ resolution, facial elements were detected and localized in terms of width, height and orientation, the faces were placed centrally with eyes aligned horizontally relative to the image borders, were rescaled so that there were approximately 100 pixels between the eyes, and so that each face occupied roughly 60\% of the total image area. A generic blurred background constituting 40\% of the square area was generated for the images.

\subsection{Pretrained generator models}
\begin{figure}[h]
\begin{center}
\framebox{
\includegraphics[width=15cm]{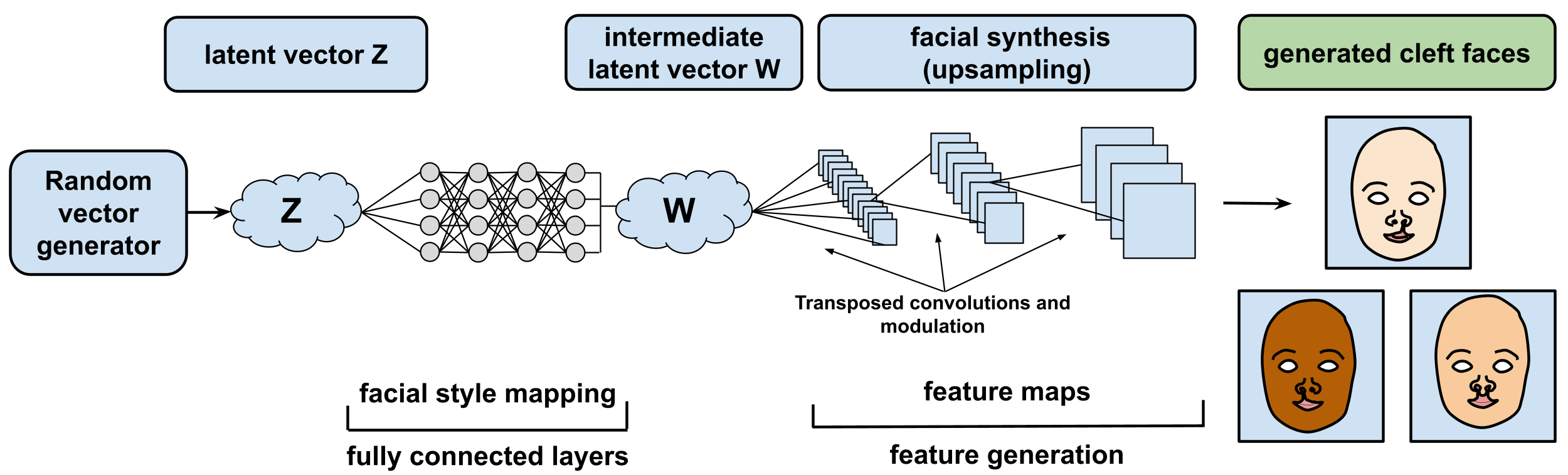}}
\end{center}
\caption{\label{img:CleftGAN}The basic structure of the StyleGAN generator consists of a mapping network encoding object features, and a synthesis network gradually constructing images based on the styles represented in the latent vector w and additional noise elements. }
\end{figure}

The StyleGAN facial base model was initially trained using the Flickr-Faces-High-Quality (FFHQ) dataset, which consists of 70,000 PNG images at 1024×1024 resolution with a broad variation in expression, age, ethnicity and image background \cite{ffhqDataset}. The dataset was carefully curated to include only high quality images with natural appearance. Many of the images in the dataset underwent some preprocessing steps to standardize the resolution and alignment of facial features. In the present study, StyleGAN2-ADA, StyleGAN3-t, and StyleGAN3-r were applied and tested. All three StyleGAN variants share the same basic architecture. A simplified illustration of the generator side of the model is shown in Figure \ref{img:CleftGAN} and follows the common design of style-based GANs in general \cite{melnik2022face}. The input to the generator network is a 512 dimensional latent vector z that semantically encodes different aspects of a sample from the  represented domain. The latent vector z is then transformed into an intermediate latent vector w which provides better latent separation between the object’s latent features. Subsequently, a series of upsampling convolutions are applied to w during the synthesis phase, enabling the gradual generation of a facial image. An analysis of the particular advantages and disadvantages of StyleGAN2 and StyleGAN3 is provided in the Discussion below.

\subsubsection{Data Augmentation}
A convolutional neural network requires a large training dataset to facilitate optimal tuning of the internal parameters of the network. However, the development of CleftGAN was undertaken to specifically address the scarcity of available cleft images. Therefore, data augmentation methods (e.g. color and geometric transformation, pixel blitting, image corruption) were required to expand our original set of training images by introducing different semantic-preserving variations. Dozens of different combinations of augmentation maneuvers were tested for optimal performance (see Performance Measures, below). 
 
In addition to testing our transfer learning protocols in stages with progressively larger image datasets, and on 3 versions of StyleGAN, we also compared four different augmentation regimens based on the recommendations of Karras et al \cite{karras2020training}: (1) training without any augmentation, (2) color augmentation, applying variation in brightness, contrast, hue and saturation, and random luma flip (“c”), (3) blitting/geometrical transformation such as horizontal flip, 90 degree rotation, random isotropic/anisotropic scaling, and integer/fractional translation (“bg”), and (4) a random combination of the previously mentioned augmentation operations (“bgc”).. The generation of transformed images enabled us to amplify our dataset and rigorously tune the StyleGAN models under analysis.

\subsubsection{Performance measures}
Three methods were used to evaluate CleftGAN’s ability to produce a high-quality facsimile of an image depicting a cleft lip: Frechet Inception Distance \cite{heusel2017gans} (FID), Perceptual Path Length \cite{karras2019style} (PPL), and a new scale we call the Divergent Index of Severity Histograms (DISH). Better performance is reflected by lower values for all three measures, indicating greater similarity between the distribution of the original dataset and the distribution of the generated samples. 

The FID is a metric commonly used to compare the quality and diversity of two image sample sets, typically generated by a GAN. The FID of two sets X1 and X2 are expressed in the following equation:

\begin{equation}
FID(x_1,x_2)=||\mu_1-\mu_2||_2^2+Tr(\Sigma_1+\Sigma_2-2(\Sigma_1\Sigma_2)^{\frac{1}{2}})
\end{equation}

Where $\mu_1$, $\mu_2$, $\Sigma_1$ and $\Sigma_2$ are the mean vectors and covariance matrices for the embedded versions of sets $X_1$ and $X_2$, respectively. The embedding is calculated using the VGG16 convolutional neural network \cite{simonyan2014very} as a feature extractor. 

Like any metric, the FID will have greater statistical significance when considering larger size datasets. Moreover, it is generally more reliable when comparing two sets of roughly equal size since one disproportionately larger set can introduce bias in the metric. In the case of CleftGAN, the 514 image training set was small compared to the number of generated samples, so the reliability of the FID measure alone might be questioned. We thus strengthened our performance evaluation by adding additional metrics.

The smoothness of a GAN generator G can be measured using PPL.The PPL metric calculates the mean of the perceptual distance between two images generated from two very close and random latent vectors lying in the linear path between two endpoints in the latent space of G. The PPL metric can be expressed in the following equation:

\begin{equation}\label{eqn:DISH}
S=\mathbb{E}[\frac{1}{\epsilon^2}d(G(slerp(\textbf{z}_1,\textbf{z}_2;t)),G(slerp(\textbf{z}_1,\textbf{z}_2;t+\epsilon)))],
\end{equation}

where $z_1$ and $z_2$ are two endpoints in the latent space. slerp is the spherical linear interpolation, t is between $z_1$ and $z_2$, and $\epsilon$ is a very small step size in the latent space. $G$ is the generator network under analysis, and the function $d$ is the perceptual distance measure, which commonly is expressed using the Learned Perceptual Image Patch Similarity (LPIPS) \cite{zhang2018unreasonable} distance. The sum of the perceptual distances has to be linearly proportional to the length of the path between the two points in the latent space, when progressively moving in the interpolation path.  

DISH is the third metric that was used to evaluate the quality of the CleftGAN images. This measure calculates the Jensen-Shannon (JS) divergence\cite{lin1991divergence} between the histograms of image severity ratings for both the real and the generated faces. Given two finite distributions p and q with corresponding elements $x_1,..., x_n$ and $y_1,...y_n$. The JS divergence between p and q is expressed in the following equation:

\begin{equation}\label{eqn:JS}
JS(x,y)=\frac{KL(x||z)+KL(y||z)}{2}=\frac{1}{2}\sum^n_{i=1}[x_i ln(\frac{x_i}{z_i})+y_i ln(\frac{y_i}{z_i})]
\end{equation}
where $z_i=\frac{(x_i+y_i)}{2}$ and $JS(x,y)=JS(y,x)$.\\

To calculate the DISH metric, the following series of operations are applied:

\begin{itemize}
    \item Generate N fake samples from generator G
    \item Calculate the severity index  for each generated face using the method described in \cite{hayajneh2023unsupervised}. 
    \item Calculate the severity index  for each real sample using the same method. 
    \item Build the histogram of severity indexes for the fake samples and another one of the real ones.
    \item Compare the two histograms by using the JS divergence equation in 3.
\end{itemize}

The JS divergence metric is a bounded measure which expresses the relative entropy between two distributions and is based on the unbounded Kullback–Leibler (KL) divergence measure.

\subsection{Training setup}
All training was done on an Intel Xeon Gold 6140 CPU 2.3 GHz with Nvidia Tesla V100 GPU (Nvidia Corp., Santa Clara, CA, USA). The training code was written on Python 3.6 using Pytorch, OpenCV, Scipy and Skimage packages. Hyperparameter tuning was carried out using a batch size of 10 images and training termination was reached when no further improvement in performance (FID, PPL, and DISH) was appreciated. Images were sampled with replacement from the original 514 image dataset and  were divided into groups (or “ticks”) of 1000. Progress was evaluated following the completion of every 40 ticks. No other hyperparameters were tuned other than the augmentation configurations.

\subsection{Results and Discussion}

\begin{figure}[h]
\begin{center}
\framebox{
\includegraphics[width=15cm]{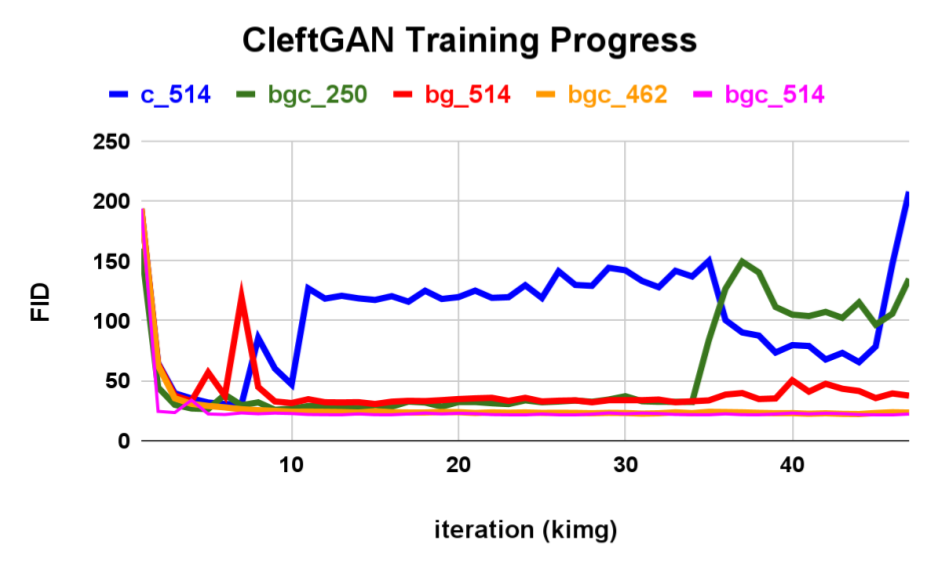}
}
\end{center}
\caption{\label{img:CleftGAN_training_progress} Training progress for three different experiments until 50k images. Bgc\_261 in the training with 261 samples. Color is the training with the whole 514 samples but utilizing the color augmentation only. The red curve demonstrates the training progress for the dataset of 514 samples using all the geometrical and the color augmentations.  }
\end{figure}

Figure \ref{img:CleftGAN_training_progress} shows FID measures during the training process for five different augmentation configurations for the StyleGAN2 model that were tested. It clearly shows that more the number of training samples (the red line versus the green dashed line) lead to lower and more stable FID values over training time. Also, the best set of augmentation methods is the bgc set, which uses the color and the geometric transformations of the images. 

The FID results of training different configurations of StyleGAN2 are shown in table \ref{tab:aug_FID}. Three sample sizes (N= 250, 450 and 514) in combination with different augmentation configurations (bg, c and bgc) as well as the possibility to use the transfer learning were tested. The minimum FID value was found to be 21.57 at the tick number 210 (840 kimg) with transfer learning using the bgc augmentation set. The FID with no augmentation and without transfer learning (the basic setup) was 41.95. the bg configuration generally results in lower FIDs of the CleftGAN generator the c configuration. This expresses the usefulness of utilizing some augmentation functions that can help prevent the problem of overfitting the discriminator during training, in which the generator represents samples existing in the dataset only, and the feedback from the discriminator to the generator becomes meaningless. This harms the convergence of the training process.
\begin{table*}
    \centering
    \begin{tabular}{|c|cccc|}
    \hline
        Sample size / other customizations &  \multicolumn{4}{c|}{FID scores/combination
of augmentations
 } \\ 
         & no aug & c & bg & bgc \\\hline
        250 faces (mixed ages) & 48.01 & 33.70 & 38.42 & 26.38\\
        450 faces (mixed ages) & 36.46 & 30.39 & 27.69 & 22.18 \\
        514 faces (95\% pediatric faces;
        no transfer learning) & 41.95  & 40.6 &28.84  & 23.75\\
        514 faces (95\% pediatric faces; transfer learning)& 31.20 & 26.34 & 23.89 & \textbf{21.57}\\\hline
    \end{tabular}
    \caption{FID results of training different setups of StyleGAN2 on cleft face samples to build the CleftGAN.}
    \label{tab:aug_FID}
\end{table*}

Using the bgc for augmentation and the FFHQ pretrained model for transfer learning, the three StyleGAN variants are further analyzed to find the best of them according to FID, PPL and DISH measures. Table \ref{tab:FID_PPL_DISH} shows the minimum obtained FID, PPL and DISH metrics for the 3 StyleGAN models under analysis. As can be noticed, the FID values are very close to each others, with minor superiority for StyleGAN3-t. However, the StyleGAN3-t is even lower PPL than the other architectures, which means that its generated images are more consistent and semantically smoother. Additionally, the DISH measure in the case of StyleGAN3-t is also the lowest among the other variants, meaning that the severity histogram of its generated images better aligns with the severity histogram of the real samples. 

\begin{table*}
    \centering
    \begin{tabular}{|c|ccc|}
    \hline
        Model &  \multicolumn{3}{c|}{Performance Metric} \\
         & FID & PPL & DISH \\ \hline
        
        StyleGAN2 & 21.57 & 19.50 & 0.148 \\
        StyleGAN3-r & 21.20 & 22.19 & 0.056 \\
        StyleGAN3-t & \textbf{21.03}  & \textbf{16.43} &\textbf{0.035}\\
        \hline
    \end{tabular}
    \caption{ FID, PPL and DISH performance measures to evaluate the three variants of StyleGAN to build the CleftGAN generator.}
    \label{tab:FID_PPL_DISH}
\end{table*}

Figure \ref{img:severity_histograms} demonstrates the difference in the level of alignment between the histograms for the  three StyleGAN variants against the severity histogram of the real samples. In the case of StyleGAN2-based architecture (a), the GleftGAN tends to represent repaired cleft lips more than the unrepaired that have higher severity indexes. This is due to the fact that during transfer learning, the domain of pretrained models is normal faces and the repaired cleft faces are closer in shape to normal faces. This in turn introduces a bias in the new domain portion that is dedicated for unrepaired versus repaired clefts. However, in the  plot (b), where StyleGAN3-t is used to build the CleftGAN, a better alignment between the severity histograms of the real and generated clefts, and more diversity of the severity indexes are obtained. This means that training StyleGAN3-t on cleft faces would better adjust the generator parameters to represent the cleft anomaly than the StyleGAN2-based CleftGAN model. StyleGAN3-r, produces a severity histogram that is more pulled towards lower severity indexes than the real severity indexes. This is due to the fact that the failure rate of generating high quality cleft faces is higher than other StyleGAN variants. 
\begin{figure}[h]
\begin{center}
\includegraphics[width=15.5cm]{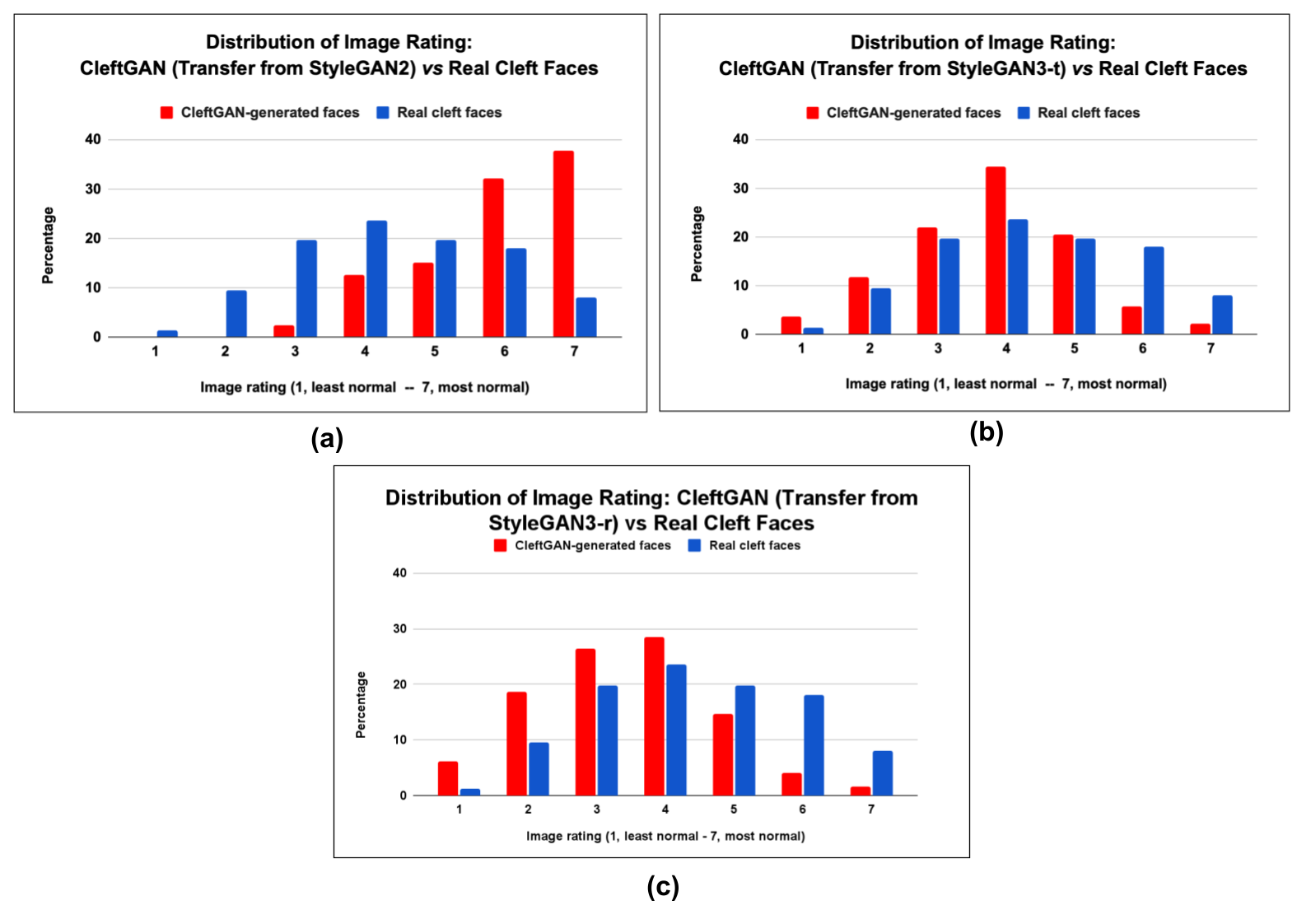}
\end{center}
\caption{\label{img:severity_histograms} The histogram of the severity indexes for different variants of CleftGAN-generated cleft faces compared to the histogram of the severity indexes for the real cleft faces. (a) StyleGAN2, (b) and StyleGAN3-t and (c) and StyleGAN3-r architectures.  }
\end{figure}
The results in this work confirm that the more the sample size available for training, the lower and better FID that can be obtained and more stable training would be done. It is better to combine and shuffle the repaired/unrepaired samples into one dataset and train the StyleGAN model. This will help in increasing the sample size and enrich the possible details of the generated  images.

The results for StyleGAN3 with translation invariance configuration showed better and lower FID than the StyleGAN2 model. However, the StyleGAN3 with rotation invariance configuration did not help in improving the results. All the StyleGAN variants feature the style-based generation of images, where the image can be modified to include specific styles of a given object class. The main difference between StyleGAN2 and StyleGAN3 is that the latter has the ability to be invariant to rotation and translation if applied in the latent space and observed in the image space (the transformation in the latent space should result in the same change in the image space). It turns out that making StyleGAN2 translation invariant improves the FID as well as PPL. This may be due to the fact that the aliasing noise coming from the generator model of StyleGAN3 is dramatically smaller that the noise coming from the StyleGAN2 generator \cite{karras2021alias} (that introduced noise does not exist in the real cleft face images). 

StyleGAN2 architecture tends to be biased to generating repaired cleft faces. For the case of StyelGAN3-t, a more stretched and aligned histogram is obtainined, representing all types of the cleft cases. StyleGAN3-r adds extra unneeded noise to the generated images so the that the histogram is pulled towards lower severy values. 

The transfer learning protocol enables the weights and biases of a neural network model to be initialized with values from a source domain (with large number of samples), then fine tune the values of the weights in the training phase from another similar domain of samples (typically with much less number of samples). This protocol works because both domains share common lower level features (curves, lines, color … etc.), and differ in the higher levels of representation, like the eyes and mouth shapes … etc. It was confirmed that transfer learning is helpful for training scenarios with low sample size \cite{oquab2014learning}. Recently, the same conclusion was obtained for the case of GANs \cite{wang2018transferring}, which aligns with what we are interested in.

Figure 5 shows two groups of generated cleft faces from CleftGAN. the first group (a) shows 25 random samples obtained by transforming the same number of randomly generated latent vectors from a uniform distribution.
The second group, (b) shows a 25 curated cleft generated faces from CleftGAN. In general, both groups are very similar to each other. However, more iamges with blue background are shown in the curated set. This means that the features of the CleftGAN high quality generated images are obtained from the high quality real samples used (the first group from the training dataset).

\begin{figure}[h]
\begin{center}
\includegraphics[width=10.5cm]{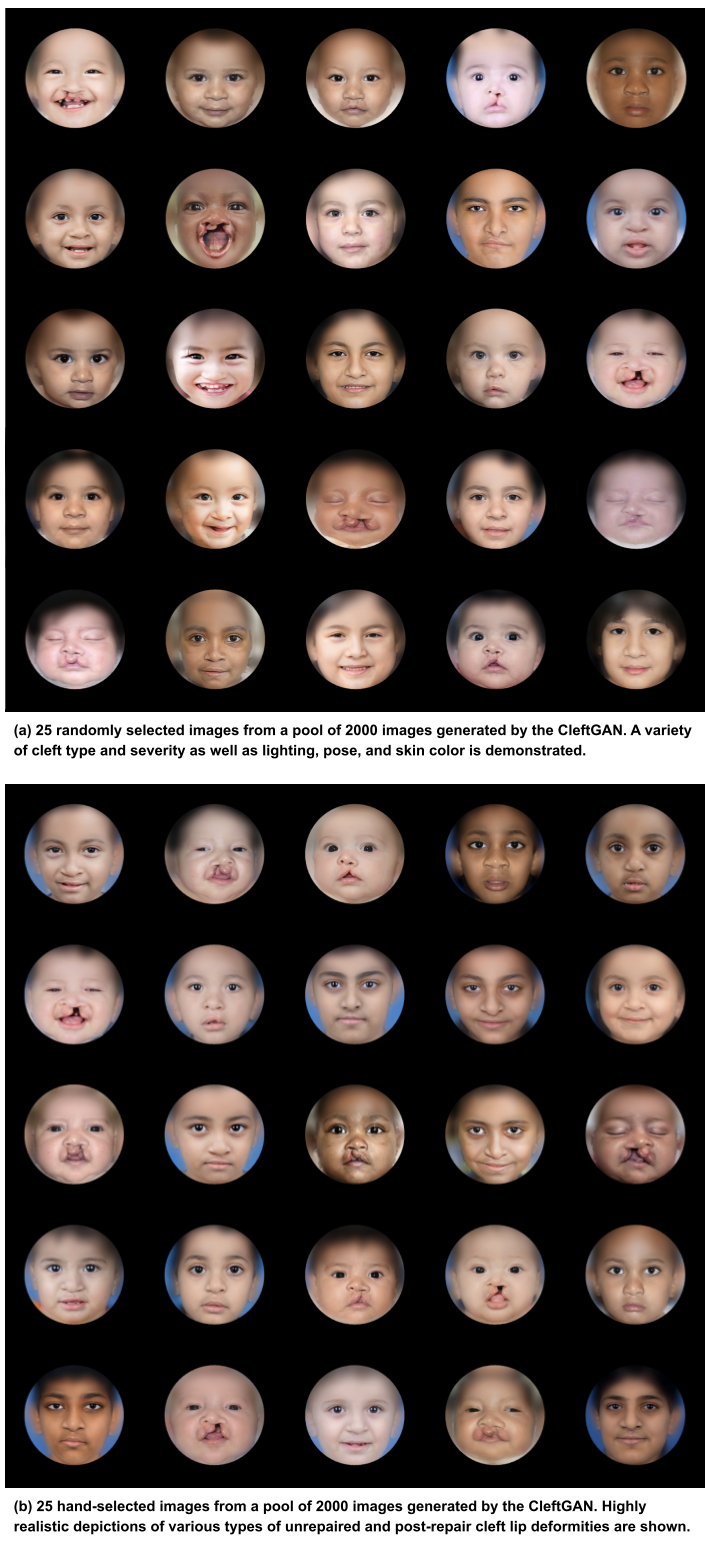}
\end{center}
\caption{\label{img:severity_histograms} Two groups of cleft faces generated from the CleftGAN generator. }
\end{figure}

\section{Conclusions}

In this work, the cleft face generator, CleftGAN, was introduced. Different GAN models in combination with augmentation methods were tested to build the CleftGAN generator. Also, the transfer learning protocol allowed to obtain better quality images that could be generated by CleftGAN. Different evaluation metrics were used to evaluate the performance of the CleftGAN model, including the FID and the PPL measures. Also, to evaluate the quality of the anomalous faces generated by CleftGAN, the severity index introduced in \cite{hayajneh2023unsupervised} were used. The results showed that the CleftGAN can generate a wide range of cleft anomaly types, which aligns - in some way - to the distribution of type of cleft anomalies in the original cleft lip dataset. 
Different type cleft and non-cleft facial anomalies can also be represented in GAN models. This would enhance the way these abnormalities could be treated, which improve the lifestyle of the patients with facial deformities. Different possible GAN architectures can be examined and evaluated to achieve better results in future work \cite{zhang2022styleswin,zhao2021improved}.

CleftGAN provides a solution to the problem of the lack of the real cleft faces and the need to have subject agreement for each sample. Furthermore, CleftGAN helps in better studying the cleft face anomaly and to better perform correction surgeries. Also, CleftGAN helps in improving the current cleft rating methods by allowing to build a large survey which includes the human ratings of a very large number of cleft cases.

The current CleftGAN generator may not be perfectly able to categorize different cleft faces in its latent space, this means that it may be hard to find a direction in the latent space that allows to adjust the severity of the cleft cases without changing the other facial features. Also the background of some collected real cleft faces is missing, which led to a cleft facial generator with no representation of the background but a blurred background instead. This motivates the need to find a way to adjust the CleftGAN to generate cleft faces with a real looking background. Another limitation of CleftGAN is that it mostly represent children's faces and it may be hard to extend it to old faces. Furthermore, the samples of the training dataset have a varying resolution and this was reflected in the CleftGAN which generates some blurred cleft faces within $1024 \times 1024$ available pixels. Finally, the cleft anomaly cases are diverse in terms of their visual appearance compared to the normal mouths. This implies that the diversity of the generated clefts may be limited to the real dataset provided.








\section*{Acknowledgments}
This publication was made possible by NPRP13S-0127-200182 from the Qatar National Research Fund (a member of Qatar Foundation). The statements made herein are solely the responsibility of the authors.

%
%
%

\bibliographystyle{acm}
\bibliography{main}

\end{document}